\documentclass[pdflatex,sn-basic,Numbered]{sn-jnl}

\usepackage{balance} 

\newcommand{\footnoteremember}[2]{%
  \refstepcounter{footnote}%
  \expandafter\xdef\csname fnnumber@#1\endcsname{\thefootnote}%
  \footnotetext{\hypertarget{#1}{#2}}%
}

\newcommand{\footnoterecall}[1]{%
  \textsuperscript{\hyperlink{#1}{\textcolor{blue}{\csname fnnumber@#1\endcsname}}}%
}

\usepackage{xcolor}
\usepackage{pgfplots}
\usepackage{tikz}
\usepackage{amssymb}
\usepackage{comment}
\usepackage{pifont}
\newcommand{\cmark}{\ding{51}}
\newcommand{\xmark}{\ding{55}}

\definecolor{bblue}{HTML}{4F81BD}
\definecolor{rred}{HTML}{C0504D}
\definecolor{ggreen}{HTML}{9BBB59}
\definecolor{ppurple}{HTML}{9F4C7C}
\definecolor{marigold}{HTML}{EAA222}
\definecolor{gunsmokegray}{HTML}{8A868E}

\usepackage{amsmath} 
\usetikzlibrary{positioning, shapes, arrows.meta} 
\usepackage{graphicx}
\usepackage{listings} 
\usepackage{xcolor} 
\usepackage{caption} 
\usepackage{subcaption}
\DeclareCaptionFormat{custom}{#1#2#3}

\usepackage{paralist}

\makeatletter
\newcommand\bigscriptsize{\@setfontsize\bigscriptsize\@viiipt\@ixpt}
\makeatother

\DeclareMathOperator{\val}{=}  
\DeclareMathOperator{\nval}{{\neq}}  

\def\happensAt{\textsf{\small happensAt}}

\def\holdsAt{\textsf{\small holdsAt}}

\def\holdsFor{\textsf{\small holdsFor}}
\def\initiatedAt{\textsf{\small initiatedAt}}
\def\terminatedAt{\textsf{\small terminatedAt}}

\def\startE{\textsf{\small start}}
\def\endE{\textsf{\small end}}

\def\unionall{\textsf{\bigscriptsize union\_all}}

\def\nbf{\textsf{\small not}}
\def\true{\textsf{\small true}}
\def\false{\textsf{\small false}}

\def\fv{$\mathit{F{=}V}$}

\def\sfluent{F}
\def\sfv{\sfluent\val V}
\def\sdfluent{F}
\def\sdfv{\sdfluent\val V}

\def\unionall{\textsf{\small union\_all}}
\def\intersectall{\textsf{\small intersect\_all}}
\def\complementall{\textsf{\small relative\_complement\_all}}

\def\atemporalconstraints{\textit{atemporal\_constraints}}
\def\intervalManipulation{\textsf{\small intervalConstruct}}

\def\rtec{RTEC}

\def\minus{{-}}

\def\qeddef{\hfill $\blacksquare$}
\def\qedex{\hfill $\lozenge$}
\def\qedprop{\hfill $\blacklozenge$}

\makeatletter
\newcommand*{\bdiv}{%
  \nonscript\mskip-\medmuskip\mkern5mu%
  \mathbin{\operator@font div}\penalty900\mkern5mu%
  \nonscript\mskip-\medmuskip
}
\makeatother

\everymath{\it}\everydisplay{\it}

\newtheoremstyle{mytheoremkr}
  {3pt}
  {3pt}
  {\normalfont}
  {0pt}
  {\bfseries}
  {.}
  { }
  {}

\theoremstyle{mytheoremkr}

\newtheorem{myexamples}{Example}

\newtheorem{mypropositions}{Proposition}

\newtheoremstyle{myproof}
  {3pt}
  {3pt}
  {\normalfont}
  {0pt}
  {\bfseries}
  {.}
  { }
  {}

\theoremstyle{myproof}

\newenvironment{mysplit}%
  {\arraycolsep 0pt \begin{array}{l}}%
  {\end{array}}

  {\begin{equation}\begin{array}{l}}%
  {\end{array}\end{equation}}

\newenvironment{logicrulenn}%
  {\begin{equation*}\begin{array}{l}}%
  {\end{array}\end{equation*}}

\allowdisplaybreaks

\usetikzlibrary{circuits.ee.IEC}

\makeatletter
\newcommand{\pushright}[1]{\ifmeasuring@#1\else\omit\hfill$\displaystyle#1$\fi\ignorespaces}
\newcommand{\pushleft}[1]{\ifmeasuring@#1\else\omit$\displaystyle#1$\hfill\fi\ignorespaces}
\makeatother

\theoremstyle{thmstyleone}%
\theoremstyle{thmstyletwo}%

\theoremstyle{thmstylethree}%
\newtheorem{definition}{Definition}%

\begin{document}

\title[Large Language Models for Executable MAS Specification Generation]{LLMs for Executable Multi-Agent System Specification Generation}


\author[1]{\fnm{Andreas} \sur{Kouvaras}}\email{a.kouvaras@unipi.gr}

\author[2]{\fnm{Periklis} \sur{Mantenoglou}}\email{periklis.mantenoglou@oru.se}

\author[1,3]{\fnm{Alexander} \sur{Artikis}}\email{a.artikis@unipi.gr}

\affil[1]{\orgname{University of Piraeus}, \orgaddress{\country{Greece}}}

\affil[2]{\orgname{Örebro University}, \orgaddress{\country{Sweden}}}

\affil[3]{\orgname{NCSR ``Demokritos''}, \orgaddress{\country{Greece}}}


\abstract{
MAS specifications express the effects of the actions of the agents and their environment, as well as other temporal phenomena, such as the intervals during which an agent may perform an action. The specification of a MAS should also be executable in order to allow for run-time monitoring.
Constructing the specification of a MAS requires formal language expertise, while machine learning techniques depend on labelled data which are rarely available.
To address these issues, we propose `genRTEC', a method that leverages pre-trained Large Language Models (LLMs) to generate executable MAS specifications, in the language of the `Run-Time Event Calculus' (RTEC), from natural language descriptions. 
genRTEC constructs MAS specifications with complex hierarchical and cyclic dependencies based only on short natural language descriptions of the concepts involved.
We present an extensive empirical evaluation of genRTEC, spanning various MAS specifications, including both a qualitative and a quantitative assessment.  
Our results demonstrate that genRTEC constructs executable MAS specifications of high predictive accuracy without compromising reasoning efficiency.}




\maketitle

\section{Introduction}

Contemporary applications are increasingly being realised in terms of multi-agent systems (MAS) \cite{DBLP:journals/corr/abs-2505-21298}. The specification of a MAS expresses the effects of the actions of the agents and their environment. Moreover, MAS are often viewed as instances of norm-governed systems \cite{DBLP:journals/ail/JonesS92}, since actuality, what is the case, and ideality, what ought to be the case, do not necessarily coincide~\cite{DBLP:journals/tosem/MinskyU00}. Therefore, a MAS specification should express the conditions in which an action is said to be permitted, prohibited and obligatory, and perhaps other more complex normative positions, as well as the enforcement strategies that handle deviations from ideality \cite{DBLP:journals/tocl/Sergot01,DBLP:journals/igpl/JonesS96}. 
In any case, MAS specifications need to explicitly represent temporal phenomena, such as the intervals during which the effects of an action persist, or the intervals during which an agent may perform some action.
Consider e.g.~an argumentation game where the agents must perform their chosen actions by specified deadlines~\cite{DBLP:journals/ai/ArtikisSP07}. Without
this feature there is no practical way of controlling the exchanges, of determining whether an agent has `spoken', because otherwise one might have to wait indefinitely for messages to arrive over the communication channels.
As another example, in voting protocols a chair of the voting procedure may be forbidden to close the ballot of a motion before a specified time period has passed~\cite{DBLP:journals/cj/PittKSA06}.
%
%
In e-commerce protocols repeated unsolicited quotes to the same consumer within a short time period may be prohibited, and may result to the (temporary) suspension of the merchant issuing them~\cite{DBLP:journals/amai/YolumS04}.

In addition to expressing temporal phenomena, the specification of a MAS should be executable, i.e.~it should be possible to compute the effects of the actions of the agents at run-time, their normative positions, and any other temporal property of the specification. This way, it would be possible to determine at run-time whether e.g.~agents comply to the specification, or whether an enforcement strategy should be applied. Such run-time monitoring should be scalable to MAS with large populations, and cater for complex specifications.

Several approaches have been proposed in the literature for developing MAS specifications. One common such approach concerns the use of the Event Calculus i.e.~a logic formalism for representing events/actions, and reasoning about their effects over time~\cite{kowalski86}.
For example, the Event Calculus has been employed for specifying e-commerce protocols~\cite{10.1093/jigpal/jzp071, chesani2009, DBLP:journals/aamas/ChesaniMMT13}, service level agreement management \cite{DBLP:journals/ijcis/FarrellSSB05, DBLP:journals/dss/PaschkeB08}, dynamic legal frameworks~\cite{DBLP:conf/icail/MarinS99, DBLP:conf/re/SharifiPALM20, DBLP:journals/sosym/ParvizimosaedSA22}, as well as authorisation policy management~\cite{DBLP:conf/policy/BandaraLR03, DBLP:conf/semweb/TontiBJMSU03, DBLP:journals/grid/ZahoorIAP22, DBLP:journals/compsec/ZahoorCAP23}.
Unlike other formal frameworks~\cite{DBLP:journals/tplp/BeckEB17,DBLP:journals/ai/BeckDE18,DBLP:journals/tplp/EiterOS19,DBLP:conf/aaai/WalegaKG19,DBLP:journals/ws/WalegaKWG23}, the Event Calculus expresses succinctly the temporal persistence of the effects of agents' actions based on the common-sense law of inertia, represents explicitly durative phenomena, and allows for hierarchical specifications paving the way towards optimised reasoning.

Constructing MAS specifications in the Event Calculus (or any other temporal logic), however, is challenging.
Domain experts may lack expertise in formal languages, while communicating the building blocks of a MAS specification to knowledge engineers can lead to information loss due to their complexity.
Moreover, automatically constructing MAS specifications via specialised machine learning techniques requires large training datasets with labelled examples, e.g.~of norm violation~\cite{DBLP:journals/pvldb/GeorgeCW16,DBLP:journals/ml/MichelioudakisA24}.
Unfortunately, such datasets are scarce, prohibiting the use of learning techniques.

To address these issues, we propose a method that employs pre-trained Large Language Models (LLMs) to construct MAS specifications in the language of the `Run-Time Event Calculus' (\rtec) \cite{DBLP:journals/tkde/ArtikisSP15}, based on natural language descriptions of the building blocks of a MAS.
%
%
%
%
%
%
%
%
\rtec\ extends the Event Calculus with optimisation techniques for temporal pattern matching over streaming data, such as streams of agent messages, outperforming  state-of-the-art systems in terms of computational efficiency~\cite{DBLP:journals/jair/TsilionisAP22, DBLP:journals/jair/MantenoglouPA25}.
RTEC, therefore, minimises the latency of the execution of a MAS specification.

%
%

The rapid evolution of LLMs has led to a class of models with the ability to elaborate on tasks by generating so-called `thinking tokens', eliciting chain-of-thought reasoning~\cite{DBLP:conf/nips/Wei0SBIXCLZ22}.
Recent works have argued that the elaboration abilities of these models do not correspond to properly structured reasoning~\cite{DBLP:conf/nips/ValmeekamMSK23,DBLP:conf/nips/DziriLSLJLWWB0H23,DBLP:conf/nips/StechlyVK24}, emphasising their limited capabilities in multi-step logical reasoning tasks~\cite{DBLP:conf/kr/IshayY023, DBLP:conf/nips/GuanVSK23, coppolillo2024llasf, DBLP:conf/ijcai/ChengLLRZL25} and in handling irrelevant information~\cite{DBLP:journals/corr/abs-2510-25626}, while other works highlight their successes in solving complicated mathematical problems~\cite{DBLP:journals/corr/abs-2501-12948}.
In any case, generating such thinking tokens is costly, in terms of both latency and monetary expenses, prohibiting the use of these models for reasoning over large, high-velocity streaming data, such as streams of agent transactions in large-scale MAS e-commerce.
On the other hand, LLMs have proven effective translators for rewriting natural language descriptions into formal language expressions~\cite{DBLP:conf/acl/YangI023,DBLP:conf/nips/GuanVSK23, DBLP:conf/emnlp/PanAWW23,DBLP:journals/pvldb/GaoWLSQDZ24, DBLP:conf/kr/IshayY023, coppolillo2024llasf, DBLP:conf/ijcnn/WangLBRZ24, DBLP:conf/aaai/Ishay025}.


We build upon this capacity for formal language generation, and present `genRTEC', i.e.~a prompting method employing pre-trained LLMs to translate natural descriptions of the building blocks of a MAS into an executable specification in the language of RTEC. 
genRTEC extends previous prompting techniques \cite{DBLP:conf/edbt/KouvarasMA25} by generalising them beyond domain-specific methods, as well as supporting the generation of specifications with cyclic dependencies, which are common in MAS.
We present the pipeline of genRTEC and the prompting practices optimising performance.
A key feature of genRTEC concerns the fact that it may be re-used, in a \emph{zero-shot manner}, to generate executable specifications for any MAS. 
We conducted an extensive, reproducible empirical evaluation of genRTEC, employing leading LLMs, and spanning MAS interaction protocols with complex hierarchical and cyclic dependencies from e-commerce, voting and argumentation. 
To stress test genRTEC further, we instructed it to generate temporal specifications for four additional application domains. 

We present a thorough quantitative assessment of the specifications constructed by genRTEC, calculating: (a) their similarity to hand-crafted counterparts acting as ground truth, thus estimating the human effort required to correct them; (b) their predictive accuracy on real and synthetic data; and (c) the computational overhead incurred when reasoning over the generated specifications as compared to the hand-crafted ones. 
Additionally, we present a qualitative analysis, identifying the main classes of errors in the generated specifications.
Our results demonstrate that genRTEC's MAS specifications achieve high predictive accuracy without compromising reasoning efficiency. The instructions to reproduce the empirical analysis are publicly available\footnoteremember{genRTEC}{\url{https://github.com/akouvaras/genRTEC}}\footnoterecall{genRTEC}.

\textbf{Running Example. }
We employ an argumentation protocol (ARG) based on the formalisation presented in~\cite{DBLP:journals/ai/ArtikisSP07}, in order to illustrate genRTEC.
There are three roles in ARG: proponent, opponent and determiner. 
We will deal with the usual case where there are three agents, one in each role.
Briefly, the proponent claims a thesis, the opponent questions this thesis, and the determiner decides whether the proponent’s thesis was successfully defended or not. 
More precisely, the argumentation commences when the proponent claims the topic of the argumentation. The protagonists --- the proponent and the opponent --- then take it in turn to perform actions, i.e.~claim, concede to, retract, or deny a proposition.
Each turn lasts for a specified time period during which the protagonist may perform several actions. After each such action the other protagonist is given an opportunity to object.
The determiner may declare the winner only at the end of the argumentation, i.e.~when the specified period for the argumentation elapses. For example, if at the end of the argumentation both the proponent and opponent have accepted the topic of the argumentation, then the determiner may only declare the proponent the winner.


\section{Related Work}\label{sec:related}

MAS specifications are temporal specifications in the sense that they express the effects of the actions of the agents and their environment, as well as other temporal phenomena, such as the intervals during which an agent has the institutional power to perform an action, and thereby create a set of institutional facts~\cite{DBLP:journals/igpl/JonesS96}, or the time until an agent should fulfill its commitments \cite{DBLP:journals/jair/ChopraVS20}.
The specification of a MAS should also be executable, i.e.~given the stream of the actions of the agents and the environment, it should be possible to compute the effects of these actions, the intervals of insitutional facts and brute facts~\cite{searle65}, and any other temporal property of the specification. To cater for large agent populations, the execution of a specification over the streaming agent actions should be performed with minimal latency.

Several approaches have been proposed in the literature for developing MAS specifications. A typical approach concerns the use of the Event Calculus, i.e.~a logic formalism for representing events/actions, and reasoning about their effects~\cite{kowalski86}.
The Event Calculus has a built-in represenetation of the common-sense law of inertia, allowing for succinct formalisations, explicitly represents durative phenomena, avoiding the issues arising from point-based semantics \cite{DBLP:journals/dss/PaschkeB08}, and supports hierarchical specifications, thus allowing the use of caching techniques for optimised reasoning \cite{DBLP:conf/time/CervesatoM00}. 
There are several works that use the Event Calculus for specifying MAS.
For instance, Symboleo is a formal specification language for smart contracts that employs the Event Calculus to define the lifecycle of contracts, as well as the obligations and the powers of agents as the system evolves~\cite{DBLP:conf/re/SharifiPALM20, DBLP:journals/sosym/ParvizimosaedSA22}.
As another example, Zahoor et al.~used the Event Calculus to model the dynamics of aggregated authorisation policies across multiple cloud providers~\cite{DBLP:journals/grid/ZahoorIAP22}, as well as to represent and reason about the state of independent authorisation policy objects in a Kubernetes cluster~\cite{DBLP:journals/compsec/ZahoorCAP23}.
%
%
The Event Calculus has also been used for monitoring temporal phenomena in mobility assistance~\cite{DBLP:journals/ijaci/BromuriUS10}, reactive and proactive health monitoring~\cite{DBLP:journals/artmed/Chaudet06, DBLP:journals/ci/KafaliRS17}, simulations with cognitive agents~\cite{DBLP:conf/ictai/ShahidOS23}, and simulations of biological feedback loops~\cite{Srinivasan2021}.

The Run-Time Event Calculus (\rtec) extends the Event Calculus with optimisation techniques, such as windowing and incremental caching, for temporal pattern matching over streaming data. It has been shown that RTEC outperforms  competing systems in terms of computational efficiency~\cite{DBLP:journals/jair/MantenoglouPA25}, including s(CASP)~\cite{DBLP:journals/tplp/AriasCCG22}, Fusemate~\cite{DBLP:conf/frocos/Baumgartner21} and jREC~\cite{chittaro96, DBLP:conf/birthday/BragagliaCMMT12, DBLP:journals/artmed/FalcionelliSTMC19}.
Moreover, \rtec\ supports temporal specifications with cyclic dependencies~\cite{DBLP:conf/kr/MantenoglouPA22}, which are common in MAS specifications, and streams with delayed actions~\cite{DBLP:journals/jair/TsilionisAP22}, which are typical in distributed systems such as MAS.

The literature includes several non-Event Calculus-based frameworks for monitoring temporal phenomena over streaming data.
CORE e.g.~is an automata-based system with formal semantics and strong time complexity guarantees~\cite{DBLP:journals/pvldb/BucchiGQRV22}.
However, CORE has limited expressivity as it supports only unary relations, applied only to the last event read.
Ticker~\cite{DBLP:journals/tplp/BeckEB17} and Laser~\cite{DBLP:conf/semweb/BazoobandiBU17} are two stream reasoning frameworks that employ fragments of the LARS temporal specification language~\cite{DBLP:journals/ai/BeckDE18}.
MeTeoR~\cite{DBLP:journals/ws/WalegaKWG23,DBLP:journals/tplp/WangGWH25} is a stream reasoning system that supports a fragment of the DatalogMTL formalism~\cite{DBLP:conf/aaai/WalegaKG19}.
StreamMill~\cite{streammill} employs a minimal extension of SQL that supports stream reasoning.
Ticker and Laser support negation only for globally stratified programs, while MeTeoR and StreamMill do not support negation in temporal patterns.
In contrast to the aforementioned frameworks, RTEC supports temporal patterns with relational constraints, such as constraints between two or more agents, and negation in locally stratified logic programs~\cite{local-strat}.
Moreover, compared to the aforementioned approaches, RTEC inherits the benefits of the Event Calculus, such as the built-in representation of inertia and the explicit modelling of durative phenomena. 

genRTEC employs pre-trained Large Language Models (LLMs) to construct executable MAS specifications in the language \rtec, based on natural language descriptions of the building blocks of a MAS. This way, MAS specifications may be constructed by users without expertise in formal languages.
Moreover, unlike specialised learning techniques~\cite{DBLP:journals/tplp/CorapiRVPS11, DBLP:journals/ml/MichelioudakisA24}, 
genRTEC does not require labelled training datasets.
LLM-based methods for executable specification generation have also been proposed for other formalisms.
Ishay et al.~\cite{DBLP:conf/aaai/Ishay025} employed LLMs to generate executable specifications in the $\mathcal{BC}+$ action language~\cite{DBLP:journals/logcom/BabbL20} in order to handle planning tasks.
As opposed to the Event Calculus, $\mathcal{BC}+$ does not support an explicit representation of time, and thus cannot express e.g.~the intervals during which a property is said to have some value. Moreover, reasoning in $\mathcal{BC}+$ is not optimised for streaming data and thus is not suitable for the run-time execution of MAS specifications.  
Guan et al.~used LLMs to construct PDDL specifications for planning problems~\cite{DBLP:conf/nips/GuanVSK23}.
PDDL, however, is not suitable for monitoring MAS, as it cannot be used for reasoning over input actions.
LLMs have also been used to generate SQL queries~\cite{DBLP:journals/pvldb/GaoWLSQDZ24}, streamlining constraints in constraint satisfaction problems~\cite{streamlining_constraints}, Answer Set Programming specifications~\cite{coppolillo2024llasf, DBLP:conf/kr/IshayY023}, and logic programs~\cite{DBLP:conf/ijcnn/WangLBRZ24, DBLP:conf/acl/YangI023}.
These formalisms are not temporal, and thus cannot be used to specify MAS.

genRTEC constitutes a domain-agnostic LLM prompting technique that builds upon a previous approach, tailored for generating RTEC rules for composite maritime activities~\cite{DBLP:conf/edbt/KouvarasMA25}.
In other words, genRTEC may be used for the construction of any MAS specification. Furthermore, specification generation is achieved in a zero-shot manner, without requiring feedback.
Note also that, unlike earlier work, genRTEC can handle  specifications with cyclic dependencies.

\section{Background: Run-Time Event Calculus}\label{sec:background}

The Run-Time Event Calculus (\rtec) is a logic programming implementation of the Event Calculus \cite{kowalski86}. Below, we briefly present the syntax of the language of RTEC and its semantics following~\cite{DBLP:journals/tkde/ArtikisSP15, DBLP:journals/jair/MantenoglouPA25}. Moreover, we outline the key reasoning task of RTEC.

\subsection{Syntax}

The language of \rtec\ includes sorts for representing (a) time, (b) events, expressing the actions of the agents and their environment, and (c) fluents, i.e.~properties whose values may change over time.
\rtec\ employs a linear timeline with non-negative integer time-points.
Variables start with an upper-case letter, while predicates and constants start with a lower-case letter.
%
A fluent-value pair (FVP) \fv\ denotes that fluent $F$ has value $V$.
Boolean fluents are a special case where the possible values are $\true$ and $\false$.
Table \ref{tbl:main_predicates} summarises the main predicates of RTEC. 
$\happensAt(E, T)$ signifies that action/event $E$ occurs at time-point $T$.
$\initiatedAt(F\val V, T)$ (resp.~$\terminatedAt(F\val V, T)$) expresses that a time period during which a fluent $F$ has the value $V$ continuously is initiated (terminated) at $T$.
$\holdsAt(F\val V, T)$ states that $F$ has value $V$ at $T$, while $\holdsFor(F\val V, I)$ expresses that \fv\ holds continuously in the maximal intervals of list $I$. 

\begin{table}[t]
\renewcommand{\arraystretch}{0.9}
\setlength\tabcolsep{4.6pt}
\begin{tabular}{ll}
\hline\noalign{\smallskip}
\multicolumn{1}{c}{\textbf{Predicate}} & \multicolumn{1}{c}{\textbf{Meaning}}  \\
\noalign{\smallskip}
\hline
\noalign{\smallskip}
\happensAt$(E, T)$ & {Event $E$ occurs at time-point $T$.}  \\[5pt]

\initiatedAt$(F \val V, T)$ & {At time-point $T$, a period of time}\\
								     & {during which $F\val V$ is initiated.} \\[5pt]

\terminatedAt$(F \val V, T)$ & {At time-point $T$, a period of time} \\
									 & {during which $F\val V$ is terminated.} \\[5pt]

\holdsFor$(F \val V, I)$ & {$I$ is the list of the maximal intervals} \\
                           					  & {during which $F\val V$ holds continuously.} \\[5pt]

\holdsAt$(F \val V, T)$ & {Fluent $F$ has value $V$ at time-point $T$.} \\[5pt]

\unionall$\mathit{([J_1,\dots,J_n],\ I)}$ & $I\val (J_1\cup\ldots\cup J_n)$ \\[5pt]

\intersectall$\mathit{([J_1,\dots,J_n],\ I)}$ & $I\val (J_1\cap\ldots\cap J_n)$ \\[5pt]

\complementall$\mathit{(I',\ [J_1,\dots,J_n],\ I)}$ & $I\val I' \setminus (J_1\cup\ldots\cup J_n)$\\[5pt]
\hline
\end{tabular}
\caption{The main predicates of RTEC.}\label{tbl:main_predicates}
\end{table}

A MAS specification may be expressed as an RTEC \emph{event description}, i.e.~a set of rules defining `simple' and `statically determined' FVPs.
A simple FVP is defined using a set of $\initiatedAt$ and $\terminatedAt$ rules, and is subject to the common-sense law of inertia, i.e.~an FVP \fv\ holds at a time-point $T$, if \fv\ has been `initiated' by an event at a time-point earlier than $T$, and not `terminated' by another event in the meantime.  

\begin{myexamples}[Concession]\label{ex:concede-effects}
Recall that in ARG the protagonists --- the proponent and the opponent --- take it in turn to perform actions, i.e.~claim, concede to, retract, or deny a proposition. 
The semantics of these actions are given in terms of the premises held by the protagonists. 
For example, the proponent's claim of a proposition $Q$ may lead to an `explicit' premise about $Q$ for the proponent and an `unconfirmed' premise about $Q$ for the opponent. 
As another example, the rules below specify the effects of a concession:
\begin{align}
& \label{eq:concede-objectionable}
    \begin{mysplit} 
    \mathit{\initiatedAt(premise(Protag, Q)\val explicit, T)} \leftarrow \\ 
    \quad\mathit{\happensAt(concede(Protag, Q), T),} \\ 
    \quad\mathit{\holdsAt(premise(Protag, Q)\val unconfirmed, T),} \\ 
    \quad\mathit{\nbf\,\holdsAt(objectionable(Protag, concede, Q)\val \true, T).} 
    \end{mysplit} \\
& \label{eq:conceded-objected}
    \begin{mysplit} 
    \mathit{\initiatedAt(premise(Protag, Q)\val explicit, T)} \leftarrow \\ 
    \quad\mathit{\happensAt(concede(Protag, Q), T),} \\ \quad\mathit{\holdsAt(premise(Protag, Q)\val unconfirmed, T),} \\ \quad\mathit{\nbf\,\happensAt(objected, T).} \end{mysplit}
\end{align}
%
$\mathit{premise(Protag,Q)}$ is a simple fluent expressing the propositions $Q$ for which a protagonist $Protag$ has an explicit or unconfirmed premise. 
$\mathit{concede(Protag,Q)}$ denotes that $\mathit{Protag}$ concedes to proposition $Q$,  `$\nbf$' expresses negation-by-failure~\cite{clark78}, $\mathit{objectionable(Protag,A,Q)}$ is a fluent expressing whether action $A$ about $Q$, performed by $Protag$, is said to be `objectionable', and $objected$ is an event denoting that at least one agent has expressed an objection.
According to rules \eqref{eq:concede-objectionable} and \eqref{eq:conceded-objected}, a protagonist $\mathit{Protag}$ is said to adopt an explicit premise about a proposition $Q$ when $\mathit{Protag}$ concedes to $Q$, provided that $\mathit{Protag}$ has an unconfirmed premise about $Q$, and the concession is not objectionable or no  agent objects to it. 
We adopt the simplistic treatment of objections of \cite{DBLP:journals/ai/ArtikisSP07}, according to which an objection to an action, such a concession, takes place at the same time as the action itself. 
A more elaborate formalisation of the objection mechanism  will be considered in future work. 
%
%
 \qedex
\end{myexamples}

\begin{definition}[Syntax of Rules Defining Simple FVPs]\label{def:rule_syntax}
The $\initiatedAt(\sfv, T)$ rules concerning a simple FVP $\sfv$ have the following syntax:
\begin{logicrulenn} 
\initiatedAt\mathit{(\sfv,\ T)} \leftarrow \\
\quad  \happensAt(E_1, T)[[, \\
\quad [\nbf]\ \happensAt(E_2, T),\\
\quad \dots, \\
\quad [\nbf]\ \happensAt(E_n,\ T), \\
\quad [\nbf]\ \holdsAt(F_1\val V_1, T),\\
\quad \dots,\\
\quad [\nbf]\ \holdsAt(F_k\val V_k, T), \\
\quad \atemporalconstraints]].
\end{logicrulenn}
\noindent The first body literal of an $\initiatedAt$ rule is a positive $\happensAt$ predicate; this is followed by a possibly empty set, denoted by `$[[\ ]]$', of positive/negative $\happensAt$ and $\holdsAt$ predicates, and $\atemporalconstraints$, i.e., a conjunction of atemporal predicates expressing background knowledge. `$\nbf$' expresses negation-by-failure, while `$[\nbf]$' denotes that `\nbf' is optional.
In addition to application-specific events, $E_i$, where $i\in [1, n]$, may refer to the built-in events of RTEC, i.e.~$\startE(F'\val V')$ or $\endE(F'\val V')$, expressing, respectively, the time-points in which FVP $F'\val V'$ is initiated or terminated \cite{DBLP:journals/tkde/ArtikisSP15}.
All (head and body) predicates are evaluated at the same time-point $T$. 
%
%
%
The body of a $\terminatedAt(\sfv, T)$ rule has the same form.
\qeddef
\end{definition}

RTEC has built-in axioms ensuring that a simple fluent cannot have more than one value at any time; an initiation of fluent $F$ with value $V'$ implies the termination of $F\val V$, for all $V\nval V'$. Moreover, a fluent may not have a value at some time-point(s). It is not the same, e.g.~to initiate $F\val\false$ and to terminate $F\val\true$: the former implies, but is not implied by the latter.

\begin{figure}[t]
    \centering
    \includegraphics[width=.5\linewidth]{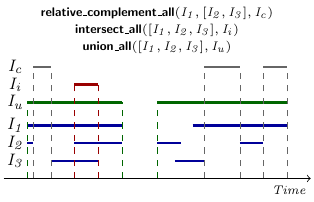}
    \caption{Interval manipulation constructs of RTEC. $I_1$, $I_2$ and $I_3$ (resp.~$I_c$, $I_i$ and $I_u$) are input (output) lists of maximal intervals.}\label{fig:constructs}
\end{figure}

A `statically determined' FVP $\sdfv$ is defined via a rule with head $\holdsFor(\sdfv, I)$.
This rule computes the maximal intervals during which $\sdfv$ holds, and it may include one or more of the interval manipulation constructs of RTEC --- see the last three items of Table \ref{tbl:main_predicates}. 
$\unionall(L, I)$ (resp.~$\intersectall(L,I)$) computes the list of maximal intervals $I$ as the union (intersection) of all lists of maximal intervals of list $L$.
%
%
$\complementall(I', L, I)$ computes the list of maximal intervals $I$ by removing from the maximal intervals of list $I'$ all interval segments included in an interval of some interval list in $L$. 
Figure \ref{fig:constructs} presents a visual illustration of the interval manipulation constructs. 

\begin{myexamples}[Objectionable concession]\label{ex:aom}
In ARG, a concession is said to be objectionable when it is `proper' but not `timely':
    \begin{align}
    & \label{eq:objectionable-concession}
    \begin{mysplit} 
    \mathit{\holdsFor(objectionable(Protag, concede, Q)\val \true, I)} \leftarrow \\ 
    \quad\mathit{\holdsFor(proper(Protag, concede, Q)\val \true, I_{1}),} \\ 
    \quad\mathit{\holdsFor(timely\val Protag, I_{2}),} \\ 
    \quad\mathit{\complementall(I_{1}, [I_{2}], I).} 
    \end{mysplit}
    \end{align}
    \noindent $\mathit{objectionable}$ is a statically determined fluent, defined in terms of the $\mathit{proper}$ and $\mathit{timely}$ fluents.
    The list of maximal intervals $I$ during which a concession by a protagonist $\mathit{Protag}$ to proposition $Q$ is said to be objectionable is computed by the relative complement of the list of maximal intervals $I_1$ during which the concession to $Q$ from $Protag$ is said to be proper, and the list of maximal intervals $I_2$ during which it is timely for $\mathit{Protag}$ to speak. See \cite{DBLP:journals/ai/ArtikisSP07} for the specification of proper and timely actions.
    \qedex
    %
\end{myexamples}

\begin{definition}[Syntax of Rules Defining Statically Determined FVPs]\label{def:rule_syntax_sdf}
The definition of a statically determined FVP $\sdfv$ is a rule with the following syntax:
\begin{logicrulenn}\label{eq:sdf-holdsFor}
\holdsFor(\sdfv,\ I_{n{+}m}) \leftarrow \\
\quad \holdsFor(F_1\val V_1,\ I_1)[[, \\
\quad \holdsFor(F_2\val V_2,\ I_2),\\
\quad \dots, \\
\quad \holdsFor(F_n\val V_n,\ I_n),\\
\quad \intervalManipulation(L_1,\ I_{n+1}),\\
\quad \dots, \\
\quad \intervalManipulation(L_m,\ I_{n+m}), \\ 
\quad \atemporalconstraints]].
\end{logicrulenn}
\noindent The first body literal of a $\holdsFor$ rule defining $\sdfv$ is a $\holdsFor$ predicate expressing the maximal intervals of an FVP other than $\sdfv$.
This is followed by a possibly empty list, denoted by `$[[\ ]]$', of $\holdsFor$ predicates, interval manipulation constructs, expressed by $\intervalManipulation$, and atemporal constraints
expressing background knowledge.  
An interval manipulation construct may be $\unionall(L_j, I_{n+j})$, $\intersectall(L_j, I_{n+j})$ or $\complementall(I_k, L_j, I_{n+j})$.
$I_k$, where $k<n+j$, is a list of maximal intervals appearing earlier in the body of the rule,
and list $\mathit{L_j}$ contains a subset of these lists.
%
%
%
%
%
The output list $I_{n{+}m}$ contains the maximal intervals during which $\sdfv$ holds continuously.
\qeddef
\end{definition}
%

%
A statically determined FVP holds as long as a Boolean combination of other FVPs is satisfied. Typically, a statically determined FVP representation leads to more efficient reasoning, but not all simple FVPs are translatable to statically determined ones~\cite{DBLP:conf/aaai/MantenoglouA25}.

\subsection{Semantics}

An event description in RTEC defines a \emph{dependency graph} expressing the relationships between its FVPs.
\begin{definition}[Dependency Graph]\label{def:dependency-graph}
\normalfont{The dependency graph of an event description is a directed graph such that:
\begin{compactenum}
 \item Each vertex denotes an FVP $F\val V$.
 \item There exists an edge $(F_j\val V_j, F_i\val V_i)$ iff there is an \initiatedAt\ or \terminatedAt\ rule for $F_i\val V_i$ having~\holdsAt$\mathit{(F_j\val V_j,\ T)}$ as one of its conditions. \hfill
 \item There exists an edge $(F_j\val V_j, F_i\val V_i)$ iff there is a \holdsFor\ rule for $F_i\val V_i$ having~\holdsFor$\mathit{(F_j\val V_j,\ T)}$ as one of its conditions. \hfill $\blacksquare$
\end{compactenum}}
\end{definition}
Based on the dependency graph, it is possible to define a function \emph{level} that maps FVPs to positive integers. 
For acyclic dependency graphs, the level of an FVP is equal to the level of its vertex.
\begin{definition}[Vertex Level]\label{def:hierarchical_level}
    Given a directed acyclic graph, the level of a vertex $v$ is equal to:
\begin{compactenum}
    \item $1$, if $v$ has no incoming edges.  
     %
    \item $n$, where $n>1$, if $v$ has at least one incoming edge from a vertex of level $\mathit{n{-}1}$, while all its other incoming edges, if any, start from vertices with level $n\minus 1$ or lower. \qeddef
\end{compactenum}
%
%
\end{definition} 
In the case that the dependency graph contains cycles, the level of each FVP is derived by first contracting the vertices of the FVPs participating in each cycle into a single vertex, and then applying Definition \ref{def:hierarchical_level} on the resulting contracted dependency graph~\cite{DBLP:conf/kr/MantenoglouPA22}.

\begin{figure}[t]
    \centering
    \includegraphics[width=0.75\linewidth]{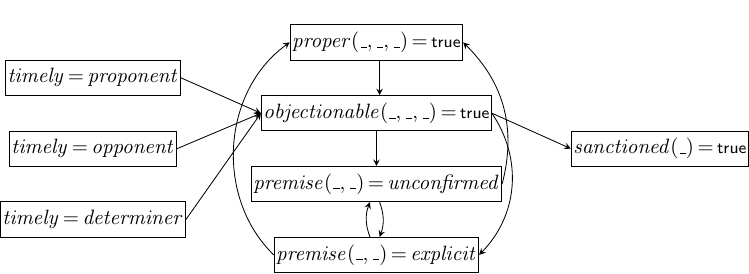}
    \caption{Dependency graph fragment of the event description for ARG. Underscores denote variables grounded on input entities, action types or agent roles. 
    }    
    \label{fig:arg_dg}
\end{figure}

\begin{myexamples}[Dependency Graph and FVP Levels] Figure \ref{fig:arg_dg} presents a dependency graph fragment of the event description for ARG. 
The FVPs with fluent $\mathit{timely}$ do not have incoming edges, as they depend only on the actions of the environment --- timeouts issued by an external clock \cite{DBLP:journals/ai/ArtikisSP07} --- and thus have level 1.
The definitions of FVPs with fluents $proper$, $objectionable$ and $premise$ exhibit cyclic dependencies.
For instance, rule \eqref{eq:concede-objectionable} stipulates that  $premise(ProtagRole, Q)\val explicit$ depends on  $objectionable(ProtagRole, concede, Q)\val\true$, while rule \eqref{eq:objectionable-concession} states that  $objectionable(ProtagRole, concede, Q)\val\true$ depends on  $proper(ProtagRole, concede, Q)\val\true$.
The event description includes additional rules stating that FVPs with fluent $proper$ depend on FVPs with fluent $premise$, completing the cycle.
To derive the levels of these FVPs, their vertices are contracted into a single vertex and the level of the resulting vertex is calculated.
This vertex has incoming edges only from vertices with level 1, and thus the levels of the FVPs taking part in this cycle are 2. 
$sanctioned$ is a fluent expressing the penalties applied to agents for performing objectionable actions. $sanctioned$ depends only on one other fluent, i.e.~$objectionable$, and thus the FVPs with fluent $sanctioned$ have level 3.    \qedex
\end{myexamples}

\begin{mypropositions}[Semantics of RTEC]\label{lemma:semantics}
An event description in RTEC is a locally stratified logic program~\cite{local-strat}. \qedprop
\end{mypropositions}
A stratification of an event description may be constructed as follows.
The first stratum contains all groundings of $\happensAt(E, T)$ atoms, expressing the actions of agents and their environment.
The remaining strata may be formed following the FVP levels in ascending order.

\subsection{Reasoning}\label{sec:RTEC-reasoning}

The key reasoning task of \rtec\ is to compute $\holdsFor(F\val V, I)$, i.e.~the list of maximal intervals $I$ during which an FVP $F\val V$ of an event description holds continuously.
For example, we may want to compute the list of maximal intervals during which an action is said to be proper in ARG, or the maximal intervals during which a protagonist is said to be sanctioned.
A statically determined FVP $\sdfv$ is defined via a rule $r$ with head $\holdsFor(\sdfv, I)$; \rtec\ derives the list of maximal intervals $I$ by evaluating the conditions of $r$.
For a simple FVP $\sfv$, \rtec\ first computes the initiation time-points of $\sfv$ by evaluating the rules with head $\initiatedAt(F\val V, T)$. Then, \rtec\ computes the termination time-points of $\sfv$ by evaluating the rules with head $\terminatedAt(F\val V, T)$ or $\initiatedAt(F\val V', T)$, where $V'\nval V$.
Subsequently, \rtec\ computes the maximal intervals of $\sfv$ by matching each initiation $T_s$ of \fv\ with the first termination $T_e$ of \fv\ after $T_s$, ignoring every intermediate initiation between $T_s$ and $T_e$.
Once the list of maximal intervals is computed, \rtec\ may  derive $\holdsAt(F\val V, T)$ by checking whether $T$ belongs to one of the maximal intervals of \fv.
RTEC includes various optimisation techniques in order to scale to high-velocity data streams. The reader is referred to \cite{DBLP:conf/kr/MantenoglouPA22, DBLP:conf/aaai/MantenoglouA25, DBLP:journals/jair/MantenoglouPA25} for details.


\section{Generating Executable Specifications}\label{sec:genRTEC}

\begin{figure}[t]
    \centering
    \includegraphics[width=0.8\linewidth]{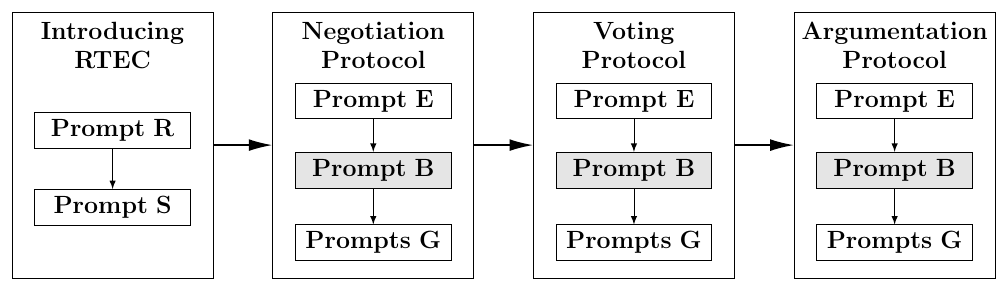}
    \caption{genRTEC: LLM prompting for executable MAS specification generation. In this illustration, genRTEC is used for the construction of the specifications of a negotation protocol, a voting protocol and an argumentation protocol. Prompt B is optional. }    
    \label{fig:mas_pipeline}
\end{figure}

Hand-crafting RTEC event descriptions requires formal language expertise, while learning such event descriptions demands large, annotated training datasets~\cite{DBLP:journals/ml/MichelioudakisA24}. To address these issues, we present genRTEC, a prompting method that leverages the power of pre-trained LLMs for constructing executable MAS specifications, i.e.~event descriptions in the language of RTEC. 

\subsection{Prompting Pipeline}

Figure \ref{fig:mas_pipeline} illustrates the workflow of genRTEC.
Initially, we introduce the language of RTEC to the LLM under consideration (in Section \ref{sec:results} we present an empirical analysis with three leading LLMs). We use prompt R to introduce the core predicates of RTEC --- see Listing \ref{lst:promptR}. Then, we issue prompt S to provide the syntax for the rules of simple and statically determined FVPs (see Definitions \ref{def:rule_syntax} and \ref{def:rule_syntax_sdf}). For each FVP representation, we provide natural language descriptions of two example time-varying properties, such as a premise in ARG, and show how these properties may be translated into RTEC rules. Through these examples, we guide the LLM in the task of mapping natural language into RTEC elements, i.e.~fluents, events, \linebreak\initiatedAt, \terminatedAt\ and \holdsFor\ rules, allowing automated rule generation. All prompts, including prompt S, are available at the repository of genRTEC\footnoterecall{genRTEC}.

\bigskip

\begin{lstlisting}[caption={Prompt R.}, label={lst:promptR}]
You are an assistant in constructing rules in the language of the Run-Time Event Calculus (RTEC), given a natural language description. The Event Calculus is a logic-based formalism for representing and reasoning about events and their effects. RTEC is a Prolog implementation of the Event Calculus, that has been optimised for stream reasoning. Below, we summarise the language of RTEC. 

Following the Prolog convention, variables start with an upper-case letter, while predicates and constants start with a lower-case letter. Each rule ends with a full-stop `.', while the head of a rule is separated from its body with `:-'. 

A fluent is a property that may have different values at different points in time. The term F=V denotes that fluent F has value V. Boolean fluents are a special case in which the possible values are `true' and `false'. 

Below are the predicates of RTEC.

RTEC - Predicate 1: happensAt(E,T)
Meaning: Event `E' occurs at time `T'.

RTEC - Predicate 2: holdsAt(F=V,T)
Meaning: The value of fluent `F' is `V' at time `T'.

RTEC - Predicate 3: holdsFor(F=V,I)
Meaning: `I' is the list of the maximal intervals during which `F=V' holds continuously.

RTEC - Predicate 4: initiatedAt(F=V,T)
Meaning: At time `T', a period of time for which `F=V' is initiated.

RTEC - Predicate 5: terminatedAt(F=V,T)
Meaning: At time `T', a period of time for which `F=V' is terminated.

RTEC - Predicate 6: union_all(L,I)
Meaning: `I' is the list of maximal intervals produced by the union of the lists of maximal intervals of list `L'.

RTEC - Predicate 7: intersect_all(L,I)
Meaning: `I' is the list of maximal intervals produced by the intersection of the lists of maximal intervals of list `L'.

RTEC - Predicate 8: relative_complement_all(In,L,I)
Meaning: `I' is the list of maximal intervals produced by the relative complement of the list of maximal intervals `In' with respect to every list of maximal intervals of list `L'.

RTEC also includes two built-in events.

Built-in event 1: start(F=V)
Meaning: Event `start(F=V)' takes place at the starting point of each maximal interval of fluent-value pair `F=V'.

Built-in event 2: end(F=V)
Meaning: Event `end(F=V)' takes place at the ending point of each maximal interval of fluent-value pair `F=V'.
\end{lstlisting}

After introducing the language of RTEC, we may instruct genRTEC to generate an executable specification for a MAS. Prompt E presents the actions of the agents and their environment, and prompt B presents the background knowledge predicates, if any. These predicates may appear in the bodies of rules --- see $atemporal\_constraints$ in Definitions \ref{def:rule_syntax} and \ref{def:rule_syntax_sdf}.
Prompts G then ask the LLM to translate a natural language description of each component of the MAS, such as the conditions in which an action is said to be proper in ARG, into a set of RTEC rules.
A key feature of genRTEC is that it is not custom to a particular MAS, but may generate executable specifications for any MAS. To achieve this, one needs to customise and repeat prompts E, B and G for each  MAS (see Figure \ref{fig:mas_pipeline}). In contrast, prompts R and S are used once and are not repeated. Note that MAS specification generation is achieved in a zero-shot manner, i.e.~it is not required to provide any feedback to the LLMs.
%

Prior to rule generation, we need to provide to the LLM the actions of the agents/environment. These actions are presented in Prompt E --- below, we present a fragment of prompt E for ARG including the actions of the agents.  

\bigskip
\begin{lstlisting}[caption={Prompt E (ARG).}, label={lst:promptE}]
You may use the following ARG actions:

ARG - Action 1: claim(Protag, Q)
Meaning: Protagonist `Protag', i.e., an agent occopying the role of proponent or opponent, claims proposition `Q'. 

ARG - Action 2: concede(Protag,Q)
Meaning: Protagonist `Protag', i.e., an agent occupying the role of proponent or opponent, concedes to proposition `Q'. 

ARG - Action 3: retract(Protag,Q)
Meaning: Protagonist `Protag', i.e., an agent occupying the role of proponent or opponent, retracts its commitment to proposition `Q'. 

ARG - Action 4: deny(Protag,Q)
Meaning: Protagonist `Protag', i.e., an agent occupying the role of proponent or opponent, denies proposition `Q'. 

ARG - Action 5: objected
Meaning: At least one agent objects to the preceding action. 
\end{lstlisting}

After prompt E, we provide a natural language description of each component of a MAS, and prompt the LLM to express it in the language of RTEC. Below, we present two such examples concerning ARG, in which an LLM is asked to generate rules expressing the effects of a `claim' action, and the specification of the conditions in which an action is said to be proper. 
The complete set of prompts is available in the repository of genRTEC\footnoterecall{genRTEC}. 

\bigskip
\begin{lstlisting}[caption={ARG -- effects of claim (Prompt G).}, label={lst:promptG-claim}]
Given the description below, provide the rules in the language of RTEC. You may use the built-in events of RTEC and the ARG actions. You may also use any of the ARG output fluents, i.e. the fluents that we will define together.

Description - `premise - claim': We aim to identify the effects of a claim. First, a protagonist adopts an explicit premise about a proposition Q, when this protagonist claims Q, and does not have an unconfirmed premise about Q. Second, a protagonist adopts an unconfirmed premise about proposition Q, when this protagonist does not have an explicit premise about Q and the other protagonist claims  Q. The aforementioned effects are realised provided that (1) the claim is not objectionable, or (2) no agent objects to the claim. The conditions in which an action is said to be `objectionable' will be presented later.
\end{lstlisting}

\bigskip 
\begin{lstlisting}[caption={ARG -- proper actions (Prompt G).}, label={lst:promptG-proper}]
Given the description below, provide the rules in the language of RTEC. You may use the built-in events of RTEC and the ARG actions. You may also use any of the ARG output fluents, i.e. the fluents that we will define together.

Description - `proper': We aim to identify the maximal intervals during which an action is considered proper for a protagonist. First, a claim of a proposition Q by a protagonist is considered proper as long as this protagonist has neither an explicit nor an unconfirmed premise about Q. Second, a concession to a proposition Q by a protagonist is considered proper as long as this protagonist has an unconfirmed premise about Q. Third, a retraction of a proposition Q by a protagonist is considered proper as long as this protagonist has an explicit premise about Q. Fourth, a denial of a proposition Q by a protagonist is considered proper as long as this protagonist has an unconfirmed premise about Q.
\end{lstlisting}

\bigskip

\subsection{Optimising Performance}

Through a series of experiments, we identified  a set of best practices that optimise the performance of genRTEC.
First, we find it useful to guide the LLM to make use of the actions that have been provided with prompt E. 
Second, we instruct the LLM to take into consideration any of the fluents that have been specified so far. See the reference to `output fluents' in line 1 of Listings \ref{lst:promptG-claim} and \ref{lst:promptG-proper}. By encouraging the LLM to reuse previously constructed specifications, we can build a hierarchical knowledge base in which higher-level patterns depend on lower-level ones. Such hierarchies yield compact RTEC event descriptions, enable caching of intermediate results, and improve reasoning efficiency \cite{DBLP:journals/tkde/ArtikisSP15}.
Third, we either state explicitly the conditions in which a time-varying property starts/stops taking place (Listing \ref{lst:promptG-claim}), or state the conditions that must be satisfied while the property is in effect (Listing \ref{lst:promptG-proper}).
Fourth, in cases where there are multiple facets in the definition of a property, we employ explicit enumeration, e.g.~`(1)', `(2)', as this helps the LLM to establish the required number of related rules.

Fifth, it is important to notify early the LLMs about cyclic dependencies, such as those in ARG --- see Figure \ref{fig:arg_dg}. 
We provide, with the use of prompts G, the definitions of the building blocks of a MAS to an LLM one at a time.
In the presence of cyclic dependencies, the definition of a building block may refer to another building block that has not yet been introduced. 
Consider Figure \ref{fig:arg_dg}; the definition of $premise$ depends on $objectionable$. Therefore, when prompting an LLM for the specification of $premise$, the conditions in which an action is said to be $objectionable$ may not have yet been presented.
In such cases, LLMs tend to generate inconsistent rule-sets. To address this issue, we inform the LLM in advance that the definitions of some building blocks will be provided later. See e.g.~the last sentence in Listing~\ref{lst:promptG-claim}. 

Listings~\ref{lst:promptG-claim} and \ref{lst:promptG-proper} are quite involved, describing intricate definitions of the effects of claims and proper actions. Nevertheless, the use of the prompting practices presented above allowed genRTEC to construct `perfect' rule-sets, i.e.~specifications identical to the ground truth (in the section that follows, we will present the metrics with which we evaluate the output of genRTEC). For instance, genRTEC responded to the prompt of Listing~\ref{lst:promptG-claim} with the rule-set below:
\begin{align}
& \label{eq:claim-objectionable}
    \begin{mysplit} 
    \mathit{\initiatedAt(premise(Protag, Q)\val explicit, T)} \leftarrow \\ 
    \quad\mathit{\happensAt(claim(Protag, Q), T),} \\ 
    \quad\mathit{\nbf\, \holdsAt(premise(Protag, Q)\val unconfirmed, T),} \\ 
    \quad\mathit{\nbf\,\holdsAt(objectionable(Protag, claim, Q)\val \true, T).} 
    \end{mysplit} \\
& \label{eq:claim-objected}
    \begin{mysplit} 
    \mathit{\initiatedAt(premise(Protag, Q)\val explicit, T)} \leftarrow \\ 
    \quad\mathit{\happensAt(claim(Protag, Q), T),} \\ \quad\mathit{\nbf\, \holdsAt(premise(Protag, Q)\val unconfirmed, T),} \\ \quad\mathit{\nbf\,\happensAt(objected, T).} \end{mysplit} \\
& \label{eq:claim-Protag2-objectionable}
    \begin{mysplit} 
    \mathit{\initiatedAt(premise(Protag_2, Q)\val unconfirmed, T)} \leftarrow \\ 
    \quad\mathit{\happensAt(claim(Protag, Q), T),} \\ 
    \quad\mathit{Protag \neq Protag_2}, \\ 
    \quad\mathit{\nbf\, \holdsAt(premise(Protag_2, Q)\val explicit, T),} \\ 
    \quad\mathit{\nbf\,\holdsAt(objectionable(Protag, claim, Q)\val \true, T).} 
    \end{mysplit} \\
& \label{eq:claim-Protag2-objected}
    \begin{mysplit} 
    \mathit{\initiatedAt(premise(Protag_2, Q)\val unconfirmed, T)} \leftarrow \\ 
    \quad\mathit{\happensAt(claim(Protag, Q), T),} \\ 
    \quad\mathit{Protag \neq Protag_2}, \\ 
    \quad\mathit{\nbf\, \holdsAt(premise(Protag_2, Q)\val explicit, T),} \\ 
    \quad\mathit{\nbf\,\happensAt(objected, T).} 
    \end{mysplit}
\end{align}
Recall that $\mathit{premise(Protag,Q)}$ is a fluent expressing the propositions $Q$ for which a protagonist $Protag$, i.e.~the proponent or the opponent, has an explicit or unconfirmed premise; $\mathit{objectionable}$ is a fluent denoting whether an action is said to be objectionable, and $\mathit{objected}$ expresses the objections of agents.
$\mathit{claim(Protag,Q)}$ denotes that $\mathit{Protag}$ claims proposition $Q$. 

Similar to the prompt of Listing~\ref{lst:promptG-claim}, genRTEC was able to generate a perfect specification for the the prompt of Listing~\ref{lst:promptG-proper} --- see the rule-set below:
\begin{align}
& \label{eq:claim-proper}
    \begin{mysplit} 
    \mathit{\holdsFor(proper(Protag, claim, Q)\val \true, I)} \leftarrow \\ 
    \quad\mathit{\holdsFor(premise(Protag, Q)\val explicit, I_1),} \\ 
    \quad\mathit{\holdsFor(premise(Protag, Q)\val unconfirmed, I_2),} \\ 
    \quad\mathit{\complementall(\omega, [I_1, I_2], I).} \\ 
    \end{mysplit} \\
& \label{eq:claim-concede}
    \begin{mysplit} 
    \mathit{\holdsFor(proper(Protag, concede, Q)\val \true, I)} \leftarrow \\ 
    \quad\mathit{\holdsFor(premise(Protag, Q)\val unconfirmed, I).} \\ 
    \end{mysplit} \\
& \label{eq:claim-retract}
    \begin{mysplit} 
    \mathit{\holdsFor(proper(Protag, retract, Q)\val \true, I)} \leftarrow \\ 
    \quad\mathit{\holdsFor(premise(Protag, Q)\val explicit, I).} \\ 
    \end{mysplit} \\
& \label{eq:claim-deny}
    \begin{mysplit} 
    \mathit{\holdsFor(proper(Protag, deny, Q)\val \true, I)} \leftarrow \\ 
    \quad\mathit{\holdsFor(premise(Protag, Q)\val unconfirmed, I).} \\ 
    \end{mysplit} 
\end{align}
Recall that $\complementall(I', L, I)$ computes the list of maximal intervals $I$ by removing from the maximal intervals of list $I'$ all interval segments included in an interval of some  list in $L$. 
Placing `$\omega$' in the first argument of $\complementall$ implies that we are interested in the absolute complement of the intervals of the lists of $L$. 

Once we have generated an RTEC event description expressing a MAS specification, such as the specification of an argumentation protocol, we may proceed with another MAS by customising and repeating prompts E, (B) and G. Recall that executable specification generation for each MAS is achieved in a zero-shot manner, i.e.~no example or feedback is provided in prompts E, B and G.

\section{Empirical Evaluation}\label{sec:results}

\subsection{Experimental Setup}

\subsubsection{Applications}

We evaluated genRTEC by generating executable specifications for three MAS interaction protocols, i.e.~the Argumentation Protocol (ARG)~\cite{DBLP:journals/ai/ArtikisSP07} that we have used as a running example, a Negotiation Protocol based on NetBill (NET)~\cite{DBLP:journals/wc/SirbuT95, DBLP:journals/amai/YolumS04, 10.1093/jigpal/jzp071}, and a Voting Protocol (VP) based on the formalisations of \cite{DBLP:journals/cj/PittKSA06, DBLP:conf/kr/MantenoglouPA22}. 
NET specifies the ways in which consumers interact with merchants to buy/sell digital goods.
VP may be summarised as follows: a committee sits and the chair opens the meeting; a member proposes a motion; another member seconds the motion; the members debate the motion; the chair calls for those in favour/against to cast their vote; finally, the motion is carried, or not, according to the standing rules of the committee.
In NET, VP and ARG, the task is to generate specifications that may be used by RTEC in order to compute, among others, the maximal intervals during which agents hold normative positions, such as institutional power, permission and obligation, as well as the sanctions applied in the cases of non-conformance to obligations and performance of forbidden actions. 
In NET e.g.~a contract defines a set of normative positions for the contracting parties; in VP the chair of the voting process is forbidden to close the ballot earlier than a specified time; and in ARG the participants are sanctioned when performing objectionable actions.

To stress test our approach further, we evaluated genRTEC by generating temporal specifications for four additional applications, i.e.~Maritime Situational Awareness (MSA)~\cite{DBLP:conf/debs/PitsikalisADRCJ19}, City Transport Management (CTM)~\cite{DBLP:journals/tkde/ArtikisSP15}, Activity Recognition (ACR)~\cite{DBLP:journals/tkde/ArtikisSP15}, and conformance checking with Clinical Guidelines (CG)~\cite{10.1007/978-3-642-28115-0_20}. In MSA, we aim to generate specifications that express various types of illegal, suspicious or dangerous (autonomous) vessel activities. Given such specifications, and a stream of vessel positional signals, RTEC may identify, at run-time, instances of such activities. In CTM,  we require specifications that may be used to monitor the performance of (autonomous) public transport vehicles. In ACR, the task is to generate specifications that may be used to recognise composite activities, such as a person leaving an object unattended, given streams of surveillance video frames annotated with symbolic information. In CG, the generated specifications should allow RTEC to assess whether the actions of medical personnel adhere to the prescribed guidelines.

\subsubsection{Evaluation Metrics}

We evaluate the specifications constructed by genRTEC by means of the following three metrics.
First, we calculate the f1-score of the generated specifications, i.e.~we compare RTEC's computed maximal intervals when operating over the genRTEC-constructed specifications, against the maximal intervals obtained when reasoning over hand-crafted specifications acting as ground truth. This way, we aim to estimate the predictive accuracy of the generated specifications. 
Second, we compute the syntactic similarity of the generated specifications to the hand-crafted ones --- this metric will be presented shortly. Note that a perfect f1-score does not necessarily imply perfect syntactic similarity (although perfect syntactic similarity implies a perfect f1-score). The use of syntactic similarity allows us to estimate the human effort required for correcting the issues, if any, of the generated specifications. 
Third, we compute the reasoning time achieved by RTEC when operating over the genRTEC-constructed specifications, and compare it against the reasoning time achieved when operating over the hand-crafted specifications.

To assess the syntactic similarity of genRTEC-constructed specifications with hand-crafted counterparts, we employed the metric of~\cite{DBLP:conf/edbt/KouvarasMA25}.
The values of this metric range from 0 to 1, with higher values indicating higher similarity.
The metric calculates the similarity between two RTEC specifications, i.e., sets of RTEC rules, in a recursive manner; their similarity is defined as the average similarity between the rules they contain, following a 1-to-1 rule mapping that maximises rule similarity.
Analogously, the similarity between two RTEC rules is equal to the average similarity of their logical literals, following a similarity-maximising mapping.
The similarity between two logical literals is 0 if they use a different predicate name or have different arities.
Otherwise, their similarity is derived by comparing one-by-one their arguments, granting a perfect similarity score 1 to constant pairs with the same name and variable pairs with the same appearances in the corresponding rules.

To evaluate the efficiency of RTEC when operating over the genRTEC-generated specifications, we measured the average reasoning time of RTEC within each sliding window, i.e.~the incrementally updated, bounded portion of the stream of agent actions currently held in memory  \cite{DBLP:journals/tkde/ArtikisSP15}.

\subsubsection{Ground Truth and Datasets}

\begin{figure}[h]
\centering
\begin{subfigure}{0.4\textwidth}
    \centering
    \includegraphics[width=\linewidth]{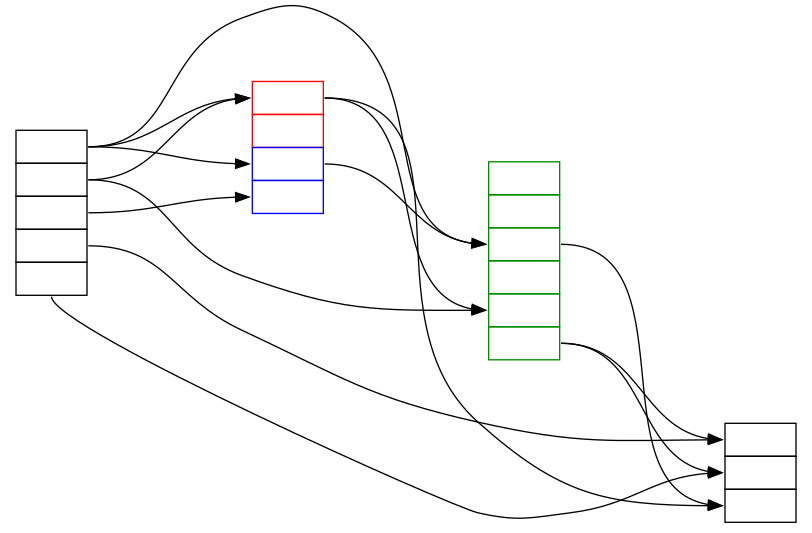}
    \caption{NET}
    \label{fig:net_graph}
\end{subfigure}
\begin{subfigure}{0.3\textwidth}
    \centering
    \includegraphics[width=\linewidth]{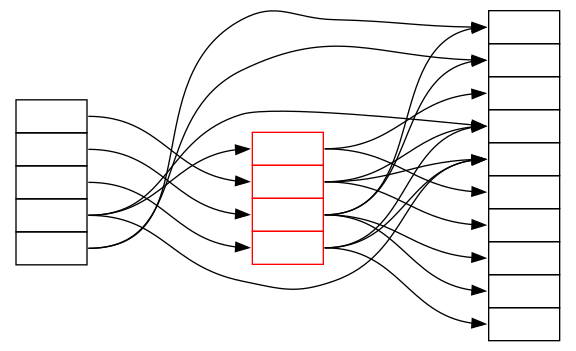}
    \caption{VP}
    \label{fig:vp_graph}
\end{subfigure}
\begin{subfigure}{0.4\textwidth}
    \centering
    \includegraphics[width=\linewidth]{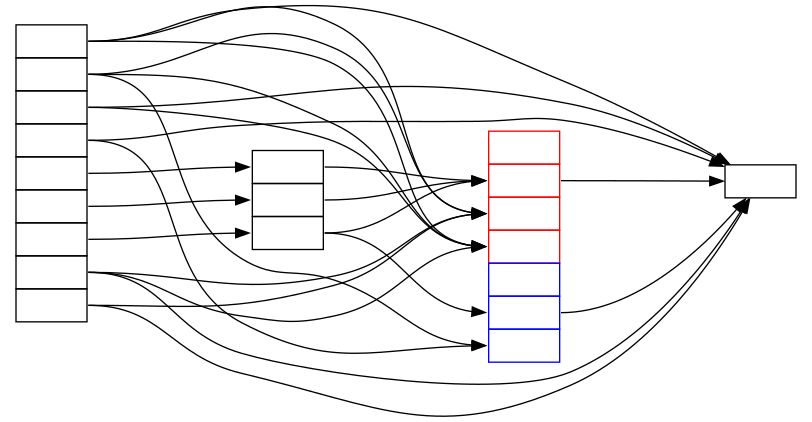}
    \caption{ARG}
    \label{fig:arg_graph}
\end{subfigure}
\caption{The dependency graphs of the hand-crafted specifications of NET, VP and ARG. To avoid clutter, we omitted the FVPs from the nodes and the edges in a cycle. 
The nodes in a cycle are coloured red, blue  or green. We grouped together the nodes of the same level, i.e.~NET has four levels, VP has three levels, and ARG has four levels. }
\label{fig:graphs_MAS}
\end{figure}

For NET, VP, MSA, ACR and CTM, there are publicly available hand-crafted event descriptions in the language of RTEC that may be used as ground truth\footnoteremember{rtec}{\url{https://github.com/aartikis/rtec}}\footnoterecall{rtec}. 
Concerning CG, some indicative Event Calculus rules are presented in \cite{10.1007/978-3-642-28115-0_20}, but these are only a small fragment of the event description, and are not entirely consistent with the syntax of the language of RTEC. 
For the employed argumentation protocol (ARG) \cite{DBLP:journals/ai/ArtikisSP07}, there is no Event Calculus specification available (the specification in \cite{DBLP:journals/ai/ArtikisSP07} is in the action language $\mathcal{C+}$ \cite{DBLP:journals/ai/GiunchigliaLLMT04}).
For these reasons, we had to construct the event descriptions for ARG and CG ourselves in order to use them as ground truth.
We have not made these event descriptions publicly available in order to make sure that LLMs will not be trained on them. 
Table \ref{tbl:app-statistics} presents the size of the hand-crafted specifications acting as ground truth. ARG e.g.~includes 24 rules, having a total of 61 conditions. Table \ref{tbl:app-statistics} also indicates whether there are cyclic dependencies in the hand-crafted specifications.
Figure \ref{fig:graphs_MAS} presents the dependency graphs of the hand-crafted event descriptions of NET, VP and ARG. The dependency graphs of the remaining applications are presented in the Appendix. 

\begin{table}[t]
\centering
\caption{Size of ground truth specifications and datasets}
\label{tbl:app-statistics}
\begin{tabular}{@{\hspace{0pt}}cccccccc@{\hspace{0.0pt}}}
\hline
 & NET & VP & ARG & MSA & CTM & ACR & CG \\ 
\hline
Rules & 15 & 15 & 24 & 32 & 23 & 9 & 19 \\
Conditions & 27 & 29 & 61 & 105 & 52 & 31 & 30 \\
Cyclic Dependencies & \cmark & \cmark & \cmark & \xmark & \xmark & \xmark & \cmark  \\
Dataset Items & 132K & 100K & 100K & 15M & 50K & 182K & N/A \\[2pt]
\hline
\end{tabular}
\end{table}

To calculate the f1-score of the genRTEC-constructed specifications and the efficiency of RTEC when reasoning over these specifications, we made use of publicly available datasets. 
In the case of NET and VP, we used datasets comprising agent interactions that were produced by synthetic data generators with realistic parameter values~\cite{DBLP:conf/kr/MantenoglouPA22}. For MSA, we employed a real dataset containing position signals emitted by vessels that sailed in the Atlantic Ocean around the port of Brest, France, between October 2015--March 2016\footnote{\url{https://zenodo.org/records/1167595}}.
In the case of CTM, we employed a synthetic dataset that was generated by simulating the operation of public transport vehicles in Helsinki, Finland \cite{DBLP:journals/ai/ArtikisSP07}.
For ACR, we used the CAVIAR benchmark dataset\footnote{\url{https://homepages.inf.ed.ac.uk/rbf/CAVIARDATA1/}}. 
All these datasets are available with the code of RTEC\footnoterecall{rtec}. Unfortunately, we could not find datasets for ARG and CG. For ARG, we created a synthetic dataset following~\cite{DBLP:journals/ai/ArtikisSP07}, while for CG we restricted attention to syntactic similarity.
Table \ref{tbl:app-statistics} presents the size of each dataset. The dataset of ARG e.g.~includes approx.~100,000 messages exchanged between the agents.

\subsubsection{Prompting}

We employed three leading LLMs, i.e.~GPT-5~\cite{openai2025gpt5}, Gemini 2.5 Pro~\cite{google2025gemini2.5} and Claude Sonnet-4~\cite{anthropic2025sonnet4.5}. For brevity, in what follows we refer to them as GPT, Gemini and Claude. We prompted each LLM, using the default hyper-parameter values, five times for each of the seven applications, and report the average values and standard deviations. 
In the experiments that follow, we used examples from MSA for prompt S of genRTEC (see Section \ref{sec:genRTEC}). We then issued customised versions of prompts E, B and G to complete the MSA specification, and subsequently generated specifications for CTM, ACR, CG, NET, VP and ARG. 
The prompts are available in the repository of genRTEC\footnoterecall{genRTEC}. The experiments were executed on a PC equipped with an Intel Core i7-1165G7 processor (2.80 GHz) and 16 GB of RAM.

\begin{figure*}[t]
\centering
\resizebox{0.82\textwidth}{!}{%
\begin{minipage}{\textwidth}
\begin{subfigure}[t]{0.24\linewidth}
    \includegraphics[width=\linewidth]{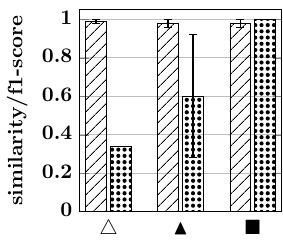}
    \caption{NET}
    \label{fig:net_scores}
\end{subfigure}%
\hfill
\begin{subfigure}[t]{0.24\linewidth}
    \includegraphics[width=\linewidth]{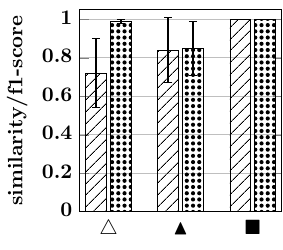}
    \caption{VP}
    \label{fig:vp_scores}
\end{subfigure}%
\hfill
\begin{subfigure}[t]{0.24\linewidth}
    \includegraphics[width=\linewidth]{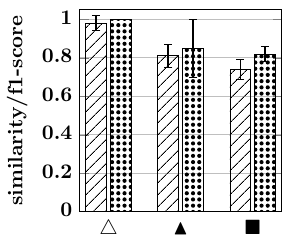}
    \caption{ARG}
    \label{fig:arg_scores}
\end{subfigure}
\hfill
\begin{subfigure}[t]{0.24\linewidth}
    \includegraphics[width=\linewidth]{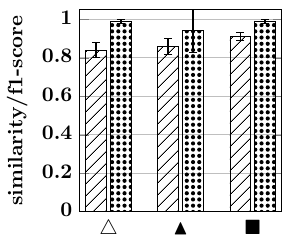}
    \caption{MSA}
    \label{fig:msa_scores}
\end{subfigure} \\ \vspace{-0.2cm}
\begin{subfigure}[t]{0.24\linewidth}
    \includegraphics[width=\linewidth]{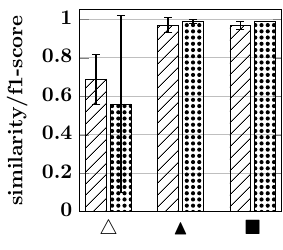}
    \caption{CTM}
    \label{fig:ctm_scores}
\end{subfigure}%
\hfill
\begin{subfigure}[t]{0.24\linewidth}
    \includegraphics[width=\linewidth]{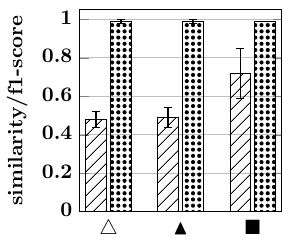}
    \caption{ACR}
    \label{fig:har_scores}
\end{subfigure}
\hfill
\begin{subfigure}[t]{0.24\linewidth}
    \includegraphics[width=\linewidth]{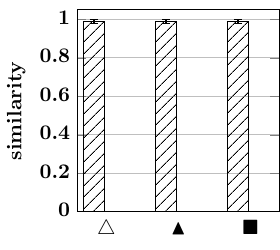}
    \caption{CG}
    \label{fig:cg_scores}
\end{subfigure}
\end{minipage}
}
\caption{Syntactic similarity (line-hatched bars) and f1-score (dotted bars) of genRTEC-constructed specifications compared to ground truth specifications. `$\triangle$', `$\blacktriangle$' and `$\blacksquare$' refer to the use of GPT, Gemini  and Claude.}
\label{fig:similarity-f1score}
\end{figure*}
\begin{figure*}[t]
\centering
\resizebox{0.62\textwidth}{!}{%
\begin{minipage}{\textwidth}
\begin{subfigure}[t]{0.32\linewidth}
    \includegraphics[width=\linewidth]{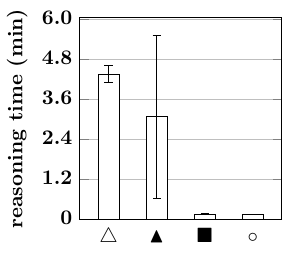}
    \caption{NET}
    \label{fig:net_time}
\end{subfigure}%
\hfill
\begin{subfigure}[t]{0.32\linewidth}
    \includegraphics[width=\linewidth]{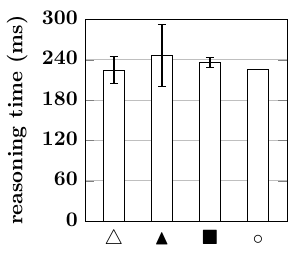}
    \caption{VP}
    \label{fig:vp_time}
\end{subfigure}%
\hfill
\begin{subfigure}[t]{0.32\linewidth}
    \includegraphics[width=\linewidth]{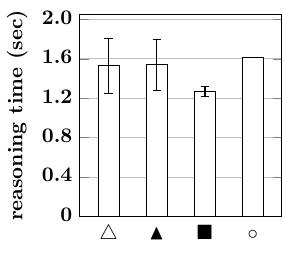}
    \caption{ARG}
    \label{fig:arg_time}
\end{subfigure}\\ \vspace{-0.1cm}
\begin{subfigure}[t]{0.32\linewidth}
    \includegraphics[width=\linewidth]{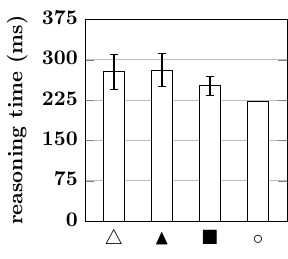}
    \caption{MSA}
    \label{fig:msa_time}
\end{subfigure}
\hfill
\begin{subfigure}[t]{0.32\linewidth}
    \includegraphics[width=\linewidth]{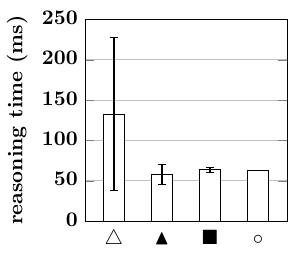}
    \caption{CTM}
    \label{fig:ctm_time}
\end{subfigure}%
\hfill
\begin{subfigure}[t]{0.32\linewidth}
    \includegraphics[width=\linewidth]{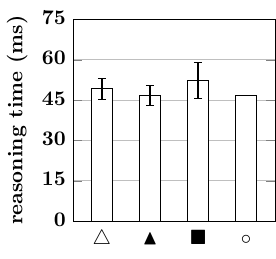}
    \caption{ACR}
    \label{fig:har_time}
\end{subfigure}%
\end{minipage}
}
\caption{Reasoning times of RTEC when operating on genRTEC-constructed specifications and ground truth specifications. `$\triangle$', `$\blacktriangle$' and `$\blacksquare$'  refer to the use of GPT, Gemini and Claude, and `$\circ$' refers to the ground truth specifications. The ranges of values in the vertical axes differ.}
\label{fig:efficiency}
\end{figure*}

\subsection{Experimental Results}

\subsubsection{Quantitative Analysis}

Figure \ref{fig:similarity-f1score} presents our experimental results with respect to syntactic similarity and f1-score, while Figure \ref{fig:efficiency} presents our results concerning efficiency. Recall that for CG there is no dataset available, and thus we can only report results for syntactic similarity. 
Figure \ref{fig:similarity-f1score} shows that the specifications generated by genRTEC match the predictive accuracy of the hand-crafted specifications, i.e.~in all applications for which we have available data, there is at least one LLM with which genRTEC achieves a(n almost) perfect f1-score. 
For example, genRTEC employing Claude achieves a perfect f1-score in NET, VP, CTM and ACR, and an f1-score just short of perfect in MSA. In ARG, genRTEC achieves a perfect f1-score when employing  GPT. 
This is a very encouraging result, especially considering that the applications under consideration require intricate specifications that include numerous rules with complicated conditions. See Table \ref{tbl:app-statistics}, Figure \ref{fig:graphs_MAS} and e.g.~rule-sets \eqref{eq:claim-objectionable}--\eqref{eq:claim-Protag2-objected} and \eqref{eq:claim-proper}--\eqref{eq:claim-deny}.
At the same time, Figure \ref{fig:efficiency} shows that, with very few exceptions, the top-performing LLMs, in terms of f1-score, generate specifications of comparable complexity with that of the ground truth specifications. Compare e.g.~the reasoning times of RTEC when operating on the NET and VP (resp.~ARG) specifications generated with the use of Claude (resp.~GPT), against the reasoning times of RTEC when operating on the corresponding ground truth specifications (see Figures \ref{fig:net_time}, \ref{fig:vp_time} and \ref{fig:arg_time}). 
This result indicates that we do not have to sacrifice predictive accuracy for efficiency, or the other way around. 

Another notable result concerns the fact that cyclic dependencies, such as those found in the specifications of NET, VP and ARG, did not compromise the performance of genRTEC. 
The prompting techniques of genRTEC (see Section \ref{sec:genRTEC}), such as explicitly informing an LLM that some of the building blocks in the body of a rule will be defined at a later stage, allowed genRTEC to capture the cyclic dependencies expressed in the hand-crafted specifications. 

The experimental results are very encouraging also in CG. Figure \ref{fig:cg_scores} shows that the generated specifications have an almost perfect syntactic similarity score. 

\subsubsection{Qualitative Analysis}

To understand the performance of genRTEC, we conducted a qualitative analysis of the generated specifications, and outline below the main issues that reduced syntactic similarity, f1-score or efficiency.

\textbf{Logical connective misrepresentation.} 
In Example \ref{ex:concede-effects} we presented the effects of conceding to a proposition in ARG. To produce rules expressing the effects of this action, we issued the following prompt:

\pagebreak 
\bigskip 
\begin{lstlisting}[caption={ARG -- effects of concession (Prompt G).}, label={lst:concede-premise}]
Description - `premise(Protag, Q)': We aim to identify the effects of a concession. A protagonist adopts an explicit premise about proposition Q, when this protagonist concedes to Q, provided that this protagonist has an unconfirmed premise about Q. The aforementioned effects are realised provided that (1) the concession is not objectionable, or (2) no agent objects to the concession. The conditions in which an action is said to be `objectionable' will be presented later. 
\end{lstlisting}
\bigskip

In some cases, genRTEC employing Claude responded with the following rule:
    \begin{align}
    & \label{eq:premise-initiatedAt-genRTEC}
    \begin{mysplit} 
    \mathit{\initiatedAt(premise(Protag, Q)\val explicit, T)} \leftarrow \\ \quad\mathit{\happensAt(concede(Protag, Q), T),} \\ 
    \quad\mathit{\holdsAt(premise(Protag, Q)\val unconfirmed, T),} \\ 
    \quad\mathit{\nbf\,\holdsAt(objectionable(Protag, concede, Q)\val \true, T),} \\
    \quad\mathit{\nbf\,\happensAt(objected, T).}
    \end{mysplit}
    \end{align}
According to rule \eqref{eq:premise-initiatedAt-genRTEC}, the effects of a concession are realised provided that the concession is not objectionable \emph{and} that no agent objects to it. 
This is an unnecessarily strict rule; e.g.~if a concession is not objectionable, then its effects should be realised, irrespective of whether some agent objected to it. 
Furthermore, according to the ground truth, i.e.~rules \eqref{eq:concede-objectionable} and \eqref{eq:conceded-objected}, the effects of a concession are realised even if the concession is objectionable, provided that no agent objects to it.
In other words, the disjunction between `objectionable concession' and any objections to it, expressed in the prompt of Listing~\ref{lst:concede-premise} --- see the penultimate sentence --- has become a conjunction in rule \eqref{eq:premise-initiatedAt-genRTEC}. 
This issue affected the syntactic similarity score of genRTEC/Claude in ARG. More importantly, it penalised the f1-score because the effects of concessions in the generated specifications were realised in much fewer cases as compared to the ground truth specifications.
Conversely, the reasoning times of RTEC when operating on the specifications generated with genRTEC/Claude were lower than the reasoning times of RTEC when operating on the hand-crafted rules, since the former specifications led to fewer/shorter intervals for $\mathit{premise}$.


\textbf{Missing rules.} We observed that, in some cases, the generated specifications included a smaller rule-set than that of the hand-crafted specifications. 
In VP e.g.~we issued the following prompt in order to produce rules recording the votes of agents: 

\bigskip
\begin{lstlisting}[caption={VP -- votes of agents (Prompt G).}, label={lst:voted}]
Description - `voted': We aim to record the way each agent has voted. Casting a vote, i.e., aye or nay, is recorded provided that the status of the motion in question is voting. An agent's vote is set to null when a new voting round begins, i.e., when the status of the motion becomes null. 
\end{lstlisting}
\bigskip 

The corresponding hand-crafted specification consists of the following rules:
\begin{align} 
    & \label{eq:voted-aye} 
    \begin{mysplit} \mathit{\initiatedAt(voted(V,M)\val aye, T)} \leftarrow \\ 
    \quad \mathit{\happensAt(vote(V,M,aye),T),} \\ 
    \quad \mathit{\holdsAt(status(M)\val voting,T).} \\ 
    \end{mysplit} \\
    & \label{eq:voted-nay} 
    \begin{mysplit} \mathit{\initiatedAt(voted(V,M)\val nay, T)} \leftarrow \\ 
    \quad \mathit{\happensAt(vote(V,M,nay),T),} \\ 
    \quad \mathit{\holdsAt(status(M)\val voting,T).} \\ 
    \end{mysplit} \\
    & \label{eq:voted-null} 
    \begin{mysplit} \mathit{\initiatedAt(voted(V,M)\val null, T)} \leftarrow \\ 
    \quad \mathit{\happensAt(\startE(status(M)\val null),T).} \\ 
    \end{mysplit} 
\end{align} 
$voted(V,M)$ is a fluent recording the vote of voter $V$ on motion $M$, $vote$ represents the act of voting, and $status$ is a fluent expressing the status of a motion. $\startE(F\val V)$ is a built-in event of RTEC taking place at the time-points in which FVP $F\val V$ is initiated (see Definition \ref{def:rule_syntax}).
genRTEC employing Gemini omitted the generation of rule \eqref{eq:voted-null}. Consequently, the votes of agents are never reset to `null', and thus may persist in future voting rounds, possibly affecting the outcome of future voting procedures. This issue affected both the syntactic similarity score and the f1-score of genRTEC/Gemini in VP. We also observed that genRTEC/Gemini sometimes produced smaller rule-sets, as compared to ground truth, in ARG.




\textbf{Fluent arity errors.} In some cases, genRTEC made errors in the arity of fluents. Consider e.g.~the specification of obligations in NET; to generate such a specification, we issued the prompt below (to simplify the presentation, we only show a fragment of the prompt): 

\bigskip
\begin{lstlisting}[caption={NET -- obligations (Prompt G).}, label={lst:obligation}]
Description - `obligation': We aim to identify whether there is a set of obligations on the contracting parties, i.e., the merchant and the consumer. First, the consumer starts being obliged to send an electronic payment order to the Intermediation Server (iServer) when the contract between the merchant and the consumer starts being in effect. [..] 
\end{lstlisting}
\bigskip 

In some cases, genRTEC generated the following rule: 
\begin{align} 
    & \label{eq:obligation-send-epo} 
    \begin{mysplit} \mathit{\initiatedAt(obl(send\_EPO(Cons, Merch, iServer, GD))\val \true, T)} \leftarrow \\ 
    \quad \mathit{\happensAt(\startE(contract(Merch, Cons, GD)\val \true),T).} \\ 
    \end{mysplit} 
\end{align}
According to rule \eqref{eq:obligation-send-epo}, a consumer $Cons$ becomes obliged to send an electronic payment order (EPO) about goods $GD$ when the contract between $Cons$ and merchant $Merch$ about $GD$ starts being in effect. Rule \eqref{eq:obligation-send-epo} is very similar to the ground truth rule. The only difference lies in the fact that the ground truth rule does not include $Merch$ in the head of the rule --- $Cons$ is obliged to send the EPO to the intermediation server $iServer$, not the merchant (see the prompt of Listing~\ref{lst:obligation}).

This discrepancy concerns the use of GPT and Gemini and affected syntactic similarity, but only very slightly (see Figure \ref{fig:net_scores}). 
On the other hand, the fluent arity discrepancy had a significant impact on the f1-score of the generated specifications, since such discrepancies did not allow us to match the FVPs of a generated specification with those of the ground truth. 
Note that, in this case, the reported f1-score is a pessimistic estimate of the predictive accuracy of the generated specifications, in the sense that the fluent arity discrepancy did not affect maximal interval computation, i.e.~the same maximal intervals were computed for the obligations of the agents when RTEC operated on the generated and hand-crafted specifications. 
The most notable effect of fluent arity discrepancy concerns computational complexity. In NET, considering all possible consumer-merchant combinations in the grounding of the fluent expressing obligations increased reasoning times substantially (see Figure \ref{fig:net_time}).

Errors in fluent arity were also observed in CTM  when genRTEC employed GPT. Syntactic similarity was affected more than in NET, because such errors were made in several fluents. Moreover, the f1-score was severely penalised and reasoning times increased. 

In CG, the syntactic similarity is slightly lower than the perfect score, because in a few experiments genRTEC added erroneously an argument to a couple of fluents. Similar to the case of NET, however, these errors will not affect maximal interval computation.


\textbf{Superfluous rule (condition) generation.} The generated specifications often include superfluous rules or rules with superfluous conditions. Consider e.g.~the fragment prompt below from VP, which was used to generate a rule-set for the status of a motion:

\bigskip 
\begin{lstlisting}[caption={VP -- status of a motion (Prompt G).}, label={lst:status}]
Description - `status': We aim to determine the status of a motion. First, the status of a motion becomes proposed when an agent proposes the motion, provided that the status of the motion is null. Second, the status changes to voting when an agent seconds the motion that was previously proposed. Third, the status becomes voted when an agent occupying the role of chair closes the ballot of the motion that was in voting status. Fourth, the status returns to null when an agent occupying the role of chair declares the outcome of the motion that was in voted status. 
\end{lstlisting}
In response to this prompt, genRTEC employing GPT sometimes generated the following rule-set: 
\begin{center}
\begin{tabular}{l @{\hspace{1.2cm}} l}  
\begin{minipage}{0.42\textwidth}
\begin{align*}
\begin{mysplit}
  \mathit{\initiatedAt(status(M)\val proposed, T)} \leftarrow \\
  \quad \mathit{\happensAt(propose(P,M), T),} \\
  \quad \mathit{\holdsAt(status(M)\val null, T).}%
\end{mysplit}
\end{align*}
\end{minipage}
&
\begin{minipage}{0.42\textwidth}
\begin{align*}
\begin{mysplit}
  \mathit{\terminatedAt(status(M)\val null, T)} \leftarrow \\
  \quad \mathit{\happensAt(propose(P,M), T),} \\
  \quad \mathit{\holdsAt(status(M)\val null, T).}
\end{mysplit}
\end{align*}
\end{minipage}
\\[1.5em]

\begin{minipage}{0.42\textwidth}
\begin{align*}
\begin{mysplit}
  \mathit{\initiatedAt(status(M)\val voting, T)} \leftarrow \\
  \quad \mathit{\happensAt(second(S,M), T),} \\
  \quad \mathit{\holdsAt(status(M)\val proposed, T).}
\end{mysplit}
\end{align*}
\end{minipage}
&
\begin{minipage}{0.42\textwidth}
\begin{align*}
\begin{mysplit}
  \mathit{\terminatedAt(status(M)\val proposed, T)} \leftarrow \\
  \quad \mathit{\happensAt(second(S,M), T),} \\
  \quad \mathit{\holdsAt(status(M)\val proposed,T).}
\end{mysplit}
\end{align*}
\end{minipage}
\\[1.5em]

\begin{minipage}{0.42\textwidth}
\begin{align*}
\begin{mysplit}
  \mathit{\initiatedAt(status(M)\val voted, T)} \leftarrow \\
  \quad \mathit{\happensAt(close\_ballot(C,M), T),} \\
  \quad \mathit{\happensAt(role\_of(C)\val chair, T),} \\
  \quad \mathit{\holdsAt(status(M)\val voting,T).}
\end{mysplit}
\end{align*}
\end{minipage}
&
\begin{minipage}{0.42\textwidth}
\begin{align*}
\begin{mysplit}
  \mathit{\terminatedAt(status(M)\val voting, T)} \leftarrow \\
  \quad \mathit{\happensAt(close\_ballot(C,M), T),} \\
  \quad \mathit{\happensAt(role\_of(C)\val chair, T),} \\
  \quad \mathit{\holdsAt(status(M)\val voting,T).}
\end{mysplit}
\end{align*}
\end{minipage}
\\[1.5em]

\begin{minipage}{0.42\textwidth}
\begin{align*}
\begin{mysplit}
  \mathit{\initiatedAt(status(M)\val null, T)} \leftarrow \\
  \quad \mathit{\happensAt(declare(C,M,\_), T),} \\
   \quad \mathit{\happensAt(role\_of(C)\val chair, T),} \\
  \quad \mathit{\holdsAt(status(M)\val voted,T).}
\end{mysplit}
\end{align*}
\end{minipage}
&
\begin{minipage}{0.42\textwidth}
\begin{align*}
\begin{mysplit}
  \mathit{\terminatedAt(status(M)\val voted, T)} \leftarrow \\
  \quad \mathit{\happensAt(declare(C,M,\_), T),} \\
   \quad \mathit{\happensAt(role\_of(C)\val chair, T),} \\
  \quad \mathit{\holdsAt(status(M)\val voted,T).}
\end{mysplit}
\end{align*}
\end{minipage}

\end{tabular}
\end{center}
$propose$, $second$, $close\_ballot$ and $declare$ express the actions of the agents; e.g.~$close\_ballot(C,M)$ states that the chair $C$ closes the ballot on motion $M$. $role\_of$ is a fluent denoting the roles of agents. 
The rules on the left column consist of the ground truth specification of $status(M)$. The rules on the right column are superfluous. 
Consider e.g.~the top two rules --- both rules have the same body literals. The presence of the top-left rule makes the top-right rule superfluous. Recall that RTEC has built-in axioms ensuring that a simple fluent cannot have more than one value at any time; an initiation of $F\val V$ implies the termination of $F\val V'$, for all $V\nval V'$ (see Section \ref{sec:background}).
This issue affected the similarity score of genRTEC when employing GPT and Gemini in VP --- see Figure \ref{fig:vp_scores}. 
Thankfully, this issue does not affect the f1-score, and has negligible effect on reasoning efficiency.
In ACR, the genRTEC-constructed specifications contain redundant conditions, which penalise syntactic similarity without affecting maximal interval computation, although they slightly increase reasoning times. The issue affected the use of all LLMs in ACR.

\section{Summary and Future Work}\label{sec:summary}

Hand-crafting MAS specifications requires knowledge of formal languages, while their automated construction via machine learning requires large and labelled training datasets, which are not always available.
To address these issues, we proposed genRTEC, i.e.~a prompting method employing pre-trained LLMs to translate natural descriptions of the building blocks of a MAS into an executable specification in the language of RTEC.
We presented the pipeline of genRTEC and the prompting practices optimising performance.
A key feature of genRTEC concerns the fact that it may be re-used, in a zero-shot manner, to generate executable specifications for any MAS. 
Our extensive empirical analysis, including leading LLMs, and spanning MAS interaction protocols with complex hierarchical and cyclic dependencies, showed that the generated MAS specifications achieve high predictive accuracy without compromising reasoning efficiency.

In the future, we aim to develop self-revision techniques \cite{DBLP:conf/aaai/Ishay025}, such as requesting from the compiler of RTEC~\cite{DBLP:conf/aaai/MantenoglouA25} to evaluate the output of an LLM --- e.g.~the dependency graph corresponding to a generated MAS specification --- and provide feedback to the LLMs.
Furthermore, we aim to fine-tune manageable versions of LLMs for avoiding the errors highlighted in our qualitative analysis.

\backmatter

\bmhead{Acknowledgements}

This work was supported partly by the EU-funded CREXDATA project (101092749), and partly by the Wallenberg AI, Autonomous Systems and Software Program (WASP) funded by the Knut and Alice Wallenberg Foundation.


\bibliography{sn-bibliography}

\pagebreak
\begin{appendices}

\section{Dependency Graphs}\label{secA1}

\begin{figure*}[htbp!]
\centering

\begin{subfigure}{0.7\textwidth}
    \centering
    \includegraphics[width=\linewidth]{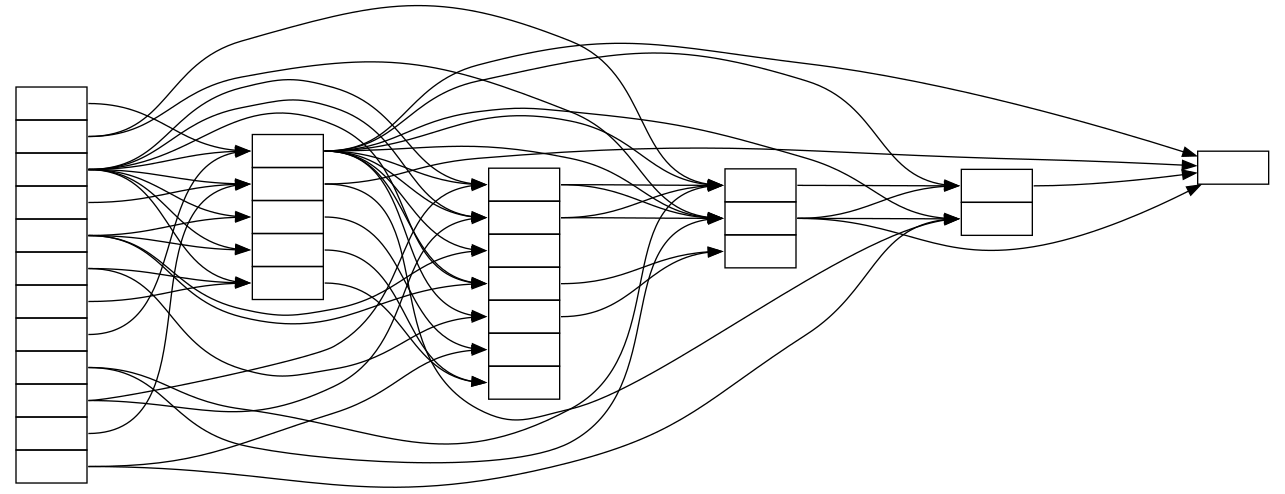}
    \caption{MSA}
    \label{fig:msa_graph}
\end{subfigure}

\begin{subfigure}{0.45\textwidth}
    \centering
    \includegraphics[width=\linewidth]{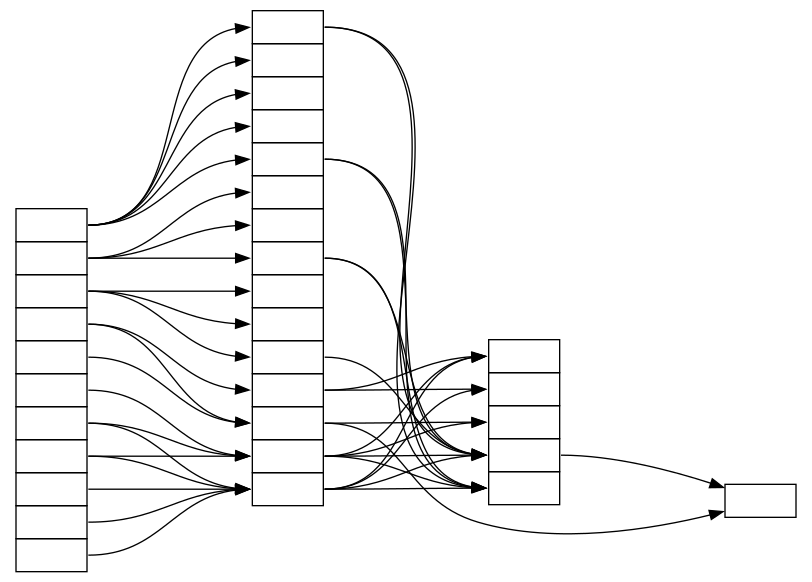}
    \caption{CTM}
    \label{fig:ctm_graph}
\end{subfigure}

\begin{subfigure}{0.7\textwidth}
    \centering
    \includegraphics[width=\linewidth]{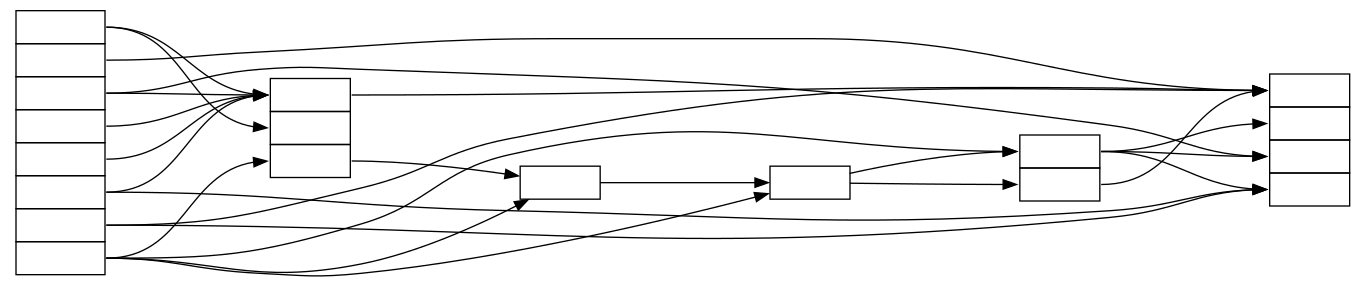}
    \caption{ACR}
    \label{fig:har_graph}
\end{subfigure}

\begin{subfigure}{0.45\textwidth}
    \centering
    \includegraphics[width=\linewidth]{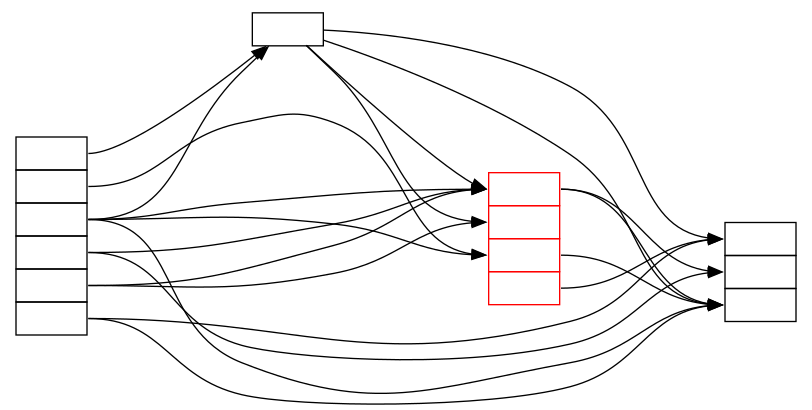}
    \caption{CG}
    \label{fig:cg_graph}
\end{subfigure}

\caption{The dependency graphs of the hand-crafted specifications of MSA, CTM, ACR and CG. The use of red colour indicates a cyclic dependency.}
\label{fig:graphs}
\end{figure*}

\end{appendices}

\end{document}